\documentclass{article}
\usepackage{spconf,amsmath,graphicx,setspace}
\usepackage{amssymb}
\usepackage{algorithm}
\usepackage{algorithmic}
\usepackage{textcomp}
\usepackage{subfigure}
\usepackage{tipa}
\usepackage{cite} 
\usepackage{url}
\usepackage{float}
\usepackage{multirow}
\usepackage{bm}

\makeatletter 
\renewcommand{\@thesubfigure}{\hskip\subfiglabelskip}
\makeatother
\title{CONTINUAL SELF-SUPERVISED LEARNING CONSIDERING MEDICAL DOMAIN KNOWLEDGE IN CHEST CT IMAGES}
\name{\begin{tabular}[t]{c}
        Ren Tasai \quad Guang Li \quad Ren Togo \quad Minghui Tang \quad Takaaki Yoshimura \quad Hiroyuki Sugimori \\
        \quad Kenji Hirata \quad Takahiro Ogawa \quad Kohsuke Kudo \quad Miki Haseyama
    \end{tabular}
\thanks{
This work was partly supported by JSPS KAKENHI Grant Numbers JP24K02942, JP23K21676, JP23K11141, and JP24K23849.
We would like to thank the departments of radiology that provided the J-MID database, including Juntendo Univ., Kyushu Univ., Keio Univ., The Univ. of Tokyo, Okayama Univ., Kyoto Univ., Osaka Univ., Tokyo Medical and Dental Univ., Hokkaido Univ., Ehime Univ., and Tokushima Univ.}}
\address{Hokkaido University}
\begin{document}
\ninept
\maketitle
%
\begin{abstract}
\end{abstract}
We propose a novel continual self-supervised learning method (CSSL) considering medical domain knowledge in chest CT images.
Our approach addresses the challenge of sequential learning by effectively capturing the relationship between previously learned knowledge and new information at different stages. 
By incorporating an enhanced DER into CSSL and maintaining both diversity and representativeness within the rehearsal buffer of DER, the risk of data interference during pretraining is reduced, enabling the model to learn more richer and robust feature representations.
In addition, we incorporate a mixup strategy and feature distillation to further enhance the model’s ability to learn meaningful representations.
We validate our method using chest CT images obtained under two different imaging conditions, demonstrating superior performance compared to state-of-the-art methods. 
\par
\begin{keywords}
Self-supervised learning, continual learning, mixup strategy, feature distillation, chest CT image.
\end{keywords}
\section{Introduction}
Self-supervised learning (SSL) has gained significant attention for its ability to reduce the annotation costs of large-scale datasets~\cite{SSL_survey2020}. 
SSL methods typically involve two phases: pretraining and fine-tuning. 
During pretraining, a model learns feature representations directly from unlabeled data. 
In the fine-tuning stage, these learned representations are refined using a smaller labeled dataset~\cite{SSL_li2022self, li2023self}. 
For large-scale medical datasets, annotations require expertise from physicians, which is both time-consuming and labor-intensive~\cite{SSL_medical_survey2020_1, SSL_medical_survey2020_2}. 
As a result, there has been active research into medical SSL methods that work with limited labeled data from various modalities such as X-rays~\cite{SSL_xray_li2023boosting, SSL_xray_li2022covid}, computed tomography (CT)~\cite{SSL_CT_covid19, SSL_CT_lung}, and magnetic resonance imaging (MRI)~\cite{SSL_multi-modality_CT_MRI, SSL_MRI}. 
These studies have shown promising results of SSL for medical image analysis.
\par
In the medical field, a variety of imaging modalities, including X-rays, CT, and MRI, are utilized. 
Due to the inherent dimensional differences between these modalities, such as 2D for X-rays and 3D for CT and MRI, recent researchers have explored SSL methods that pretrain across multiple modalities, simultaneously~\cite{medical_multi-modality_1, medical_multi-modality_2}. 
However, these studies have reported no significant improvement in classification accuracy when pretraining involves multiple modalities~\cite{medcoss}, primarily due to insufficient handling of the distinct data distributions, which causes interference during the pretraining process.
\par
Recently, continual self-supervised learning (CSSL)~\cite{medcoss} has been introduced to address data interference across modalities by considering differences in data distribution. 
CSSL preserves the diversity of data distribution during pretraining, allowing the model to maintain rich feature representations for later fine-tuning.
A major challenge in CSSL, is catastrophic forgetting, where new knowledge overwrites previously learned information~\cite{catastrophic_forgetting_1993}. 
To address this problem, a rehearsal-based approach from continual learning is employed.
This approach stores a portion of past data and features in a rehearsal buffer, enabling the model to retain and revisit prior knowledge during subsequent learning~\cite{DER, MER}.
Among these approaches, dark experience replay (DER)~\cite{DER} is particularly effective in mitigating catastrophic forgetting during continual pretraining on diverse modalities.
\par
Medical imaging relies on various domain-specific images depending on the equipment and imaging conditions~\cite{multi_window_1, multi_window_2}. 
As a result, several researches have focused on medical SSL methods that can pretrain on multiple domains simultaneously~\cite{medical_multi-domain_1, medical_multi-domain_2}. 
%
%
These domains capture the same anatomical region, whereas the modalities capture different anatomical regions. 
%
Therefore, these domains tend to have more similar data distributions compared to different modalities. 
However, prior studies often overlook differences in data distribution across domains, leading to data interference.
In CSSL across multiple domains, it is crucial to maintain not only the diversity of data distributions but also their representativeness.
%
To address this issue, we focus on capturing the relationship between past knowledge from earlier stages and future knowledge from subsequent stages. 
\par
This paper proposes a novel CSSL method that leverages medical domain knowledge in chest CT images. 
Specifically, our method incorporates an enhanced DER that ensures both diversity and representativeness in the rehearsal buffer by accounting for differences in data distributions across sequential learning stages.
The enhanced DER allows us to minimize data interference during pretraining and enables the model to learn richer feature representations across multiple domains.
Additionally, we integrate a mixup strategy and feature distillation to further improve representation learning. 
We pretrain using chest CT images obtained under two different imaging conditions and conduct evaluation on another open CT image dataset.
Through extensive experiments, our method consistently outperforms other approaches. 
\par
Our contributions are summarized as follows.
\begin{itemize}
    \item We propose a novel CSSL method that effectively addresses data distribution shifts during pretraining in chest CT images across two domains.
    \item By incorporating an enhanced DER into the CSSL method, it prevents data interference caused by the impact of different data distributions, enabling the model to acquire rich and robust feature representations.
    \item Extensive experiments show that our method outperforms state-of-the-art approaches on an open CT image dataset.
\end{itemize}
\section{Continual self-supervised learning method considering medical domain knowledge}
\begin{figure}[t]
    \centering
    \includegraphics[width=\linewidth]{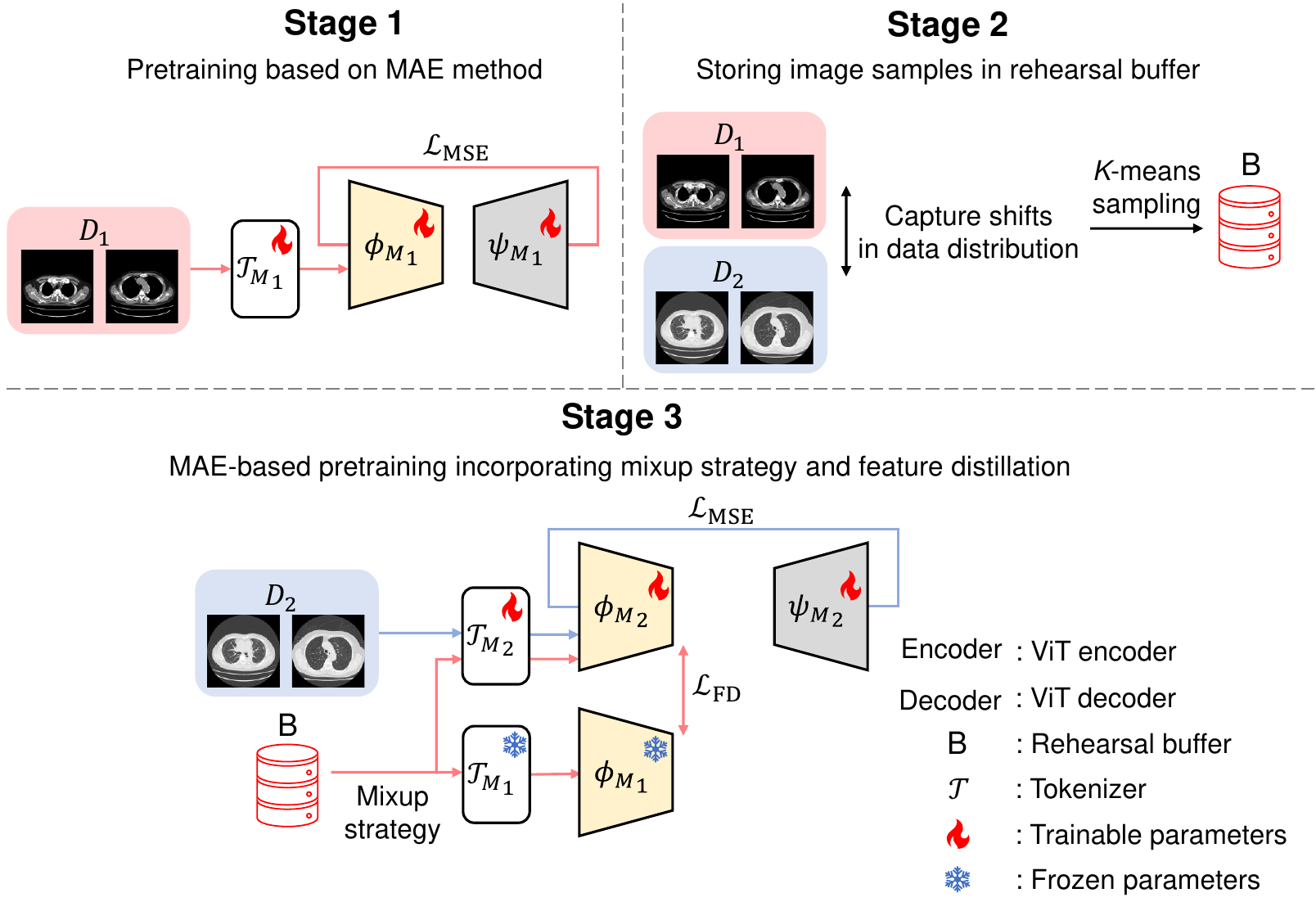} 
    \caption{Overview of the proposed CSSL method.}
    \label{fig1}
\end{figure}

The proposed method employs a three-stage CSSL approach to mitigate catastrophic forgetting and reduce data interference between two domains in chest CT images. 
In the first stage, self-supervised learning is performed using the initial dataset $D_1$ from one domain of chest CT images. 
In the second stage, selected images from $D_1$ are stored in the rehearsal buffer, ensuring that both diversity and representativeness are maintained. 
In the third stage, continual self-supervised learning is applied using the next dataset $D_2$ from another domain and the obtained rehearsal buffer. 
After completing the CSSL process, the model is fine-tuned with labeled data for downstream tasks, such as classification.
Figure~\ref{fig1} shows an overview of the proposed method.
\par
\subsection{Stage 1: Self-supervised Learning on Dataset $\boldsymbol{D_1}$}
The first stage of pretraining begins with training model $M_1$ using the first domain dataset $D_1$.
The Masked Autoencoders (MAE) method~\cite{MAE} based on masked image modeling is used to learn the input data's feature representations.
In this process, each image with $C$ channels is divided into $n$ patches of resolution $(V, V)$, and using a masking rate $r$, a subset of $m = n \times r$ patches is randomly masked.
The $n - m$ unmasked patches are converted into token sequences by the tokenizer $\mathcal{T}_{M_1}$ and are passed through the encoder $\phi_{M_1}$ to generate feature representations. 
The decoder $\psi_{M_1}$ then reconstructs the masked patches based on these representations and the embeddings of the masked patches from the original image. 
The model is trained to minimize the mean squared error (MSE) between the original masked patches $X_m$ and the reconstructed patches $Y_m$ as follows:
\begin{equation}
\mathcal{L}_{\text{MSE}} = \frac{1}{m \times V^{2}\times C} || Y_{m} - X_{m} ||^{2}_{2}.
\end{equation}
\par
At the end of the first stage, the model $M_1$ is trained, capturing the rich feature representations of the first domain dataset $D_1$.
This trained model will also be used for the third stage of CSSL, which involves both $D_1$ and the second domain dataset $D_2$.
\subsection{Stage 2: Storing Image Samples in Rehearsal Buffer}
\begin{figure}[t]
    \centering
    \includegraphics[width=7cm]{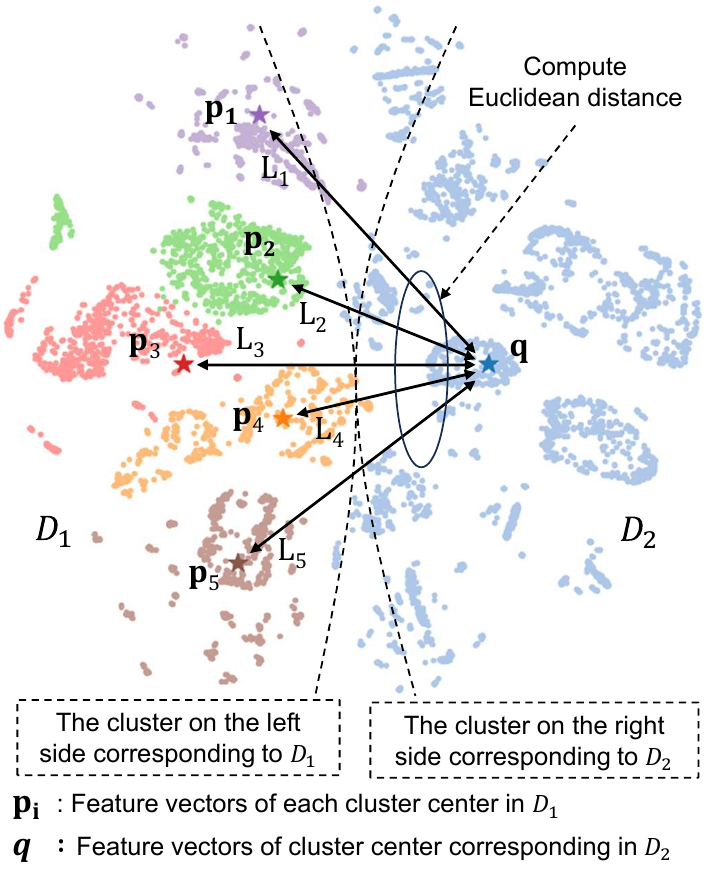}
    \caption{Overview illustration of stage 2.}
    \label{fig2}
\end{figure}

The second stage involves the selection of images stored in the rehearsal buffer, which plays a critical role in capturing shifts in data distribution between stages, helping to mitigate catastrophic forgetting. 
In the second stage, we propose an enhanced DER that uses \textit{k}-means sampling to account for the distributions of the domain datasets $D_1$ and $D_2$, ensuring that the stored samples from $D_1$ are both diverse and representative. Figure~\ref{fig2} provides an overview illustration of this stage.
\par
Let $N_1$ and $N_2$ represent the number of images in domain datasets $D_1$ and $D_2$, respectively. 
Using the parameter $\alpha$ to determine the sampling ratio, the number of clusters is set to $K = N_1 \times \alpha$.
The dataset $D_1$ is first clustered into $K$ classes based on embeddings generated by the pretrained model $M_1$, with each cluster denoted as $\mathbf{a}_i$, where $i \in \{1, 2, \dots, K\}$, and the feature vector of each cluster center represented as $\mathbf{p}_i$.
The combined dataset of $D_1$ and $D_2$ is then clustered into two classes, and the feature vector of the cluster center corresponding to dataset $D_2$ is denoted as $\mathbf{q}$. 
Next, the Euclidean distance $L_i$ from each cluster center $\mathbf{p}_i$ to $\mathbf{q}$ is computed to evaluate how close each cluster of $D_1$ is to the class center of $D_2$:
\begin{equation}
    L_i = || \textbf{p}_i - \textbf{q} ||_2.
\end{equation}
Based on these distances, clusters $\mathbf{a}_i$ are sorted in ascending order and divided into three groups: $\mathbf{G}_1$, $\mathbf{G}_2$, and $\mathbf{G}_3$. 
Image samples are selected from each group according to the parameters $\gamma_1$, $\gamma_2$, and $\gamma_3$, which determine the ratio of images retrieved from each group. 
Using the parameter $\beta$ to determine the sampling ratio, the number of images stored in the rehearsal buffer $B$ is set to $T = N_1 \times \beta$.
These samples, $T \times \gamma_1$, $T \times \gamma_2$, and $T \times \gamma_3$ are selected from the points closest to the center of each respective cluster $\mathbf{a}_i$.
The $T$ image samples $x_j$, where $j \in \{1, 2, \dots, T\}$, are stored in the rehearsal buffer $B = \{x_j \mid x_j \in D_1\}$.
The enhanced DER ensures that the rehearsal buffer $B$ used for the third stage of CSSL retains diverse and representative image samples, effectively capturing the data distributions from both $D_1$ and $D_2$.
\subsection{Stage 3: Continual Self-supervised Learning on Dataset $\boldsymbol{D_2}$}
Following the SSL pretraining of model $M_1$ in the first stage, model $M_2$ is also pretrained in the third stage using the MAE method.
In this stage, model $M_2$ is trained using both the second domain dataset $D_2$ and the image samples stored in the rehearsal buffer $B$ that obtained from the second stage.
We incorporate two critical techniques in this stage: the mixup strategy and feature distillation, both designed to enhance the model's ability to retain knowledge from the first stage while learning new representations from the second domain.
%
%
\par
The mixup strategy is applied to augment the image samples stored in the rehearsal buffer $B$, further improving the diversity of the data used for training. 
Let $S$ denote the batch size, $C$ the number of image channels, and $(H, W)$ the image resolution.
A batch of images $\mathbf{b} \in \mathbb{R}^{S \times C \times H \times W}$ is drawn from $B$, duplicated, and shuffled to create a new batch $\mathbf{b}'$. 
The mixed batch $\mathbf{b}^{\text{mix}}$ is calculated as follows:
\begin{equation}
 \mathbf{b}^{\text{mix}} = \boldsymbol{\lambda} \odot \mathbf{b} + (\mathbf{1} - \boldsymbol{\lambda}) \odot \mathbf{b}'.
\end{equation}
Here, $\boldsymbol{\lambda} \in \mathbb{R}^{S \times C \times H \times W}$ is a mask, where each element takes a random value in the range $[0, 1)$, determining the extent to which each sample in $\mathbf{b}$ is mixed with the corresponding sample in $\mathbf{b}'$. 
This mixed batch $\mathbf{b}^{\text{mix}}$ is then used to train model $M_2$, improving the representation learning process in the third stage.
%
%
\par
To further ensure that model $M_2$ retains knowledge from the first stage, feature distillation is employed. 
Specifically, the feature representations learned by $M_1$ are compared to those generated by $M_2$ during the third stage. 
The second domain dataset $D_2$ is passed only to $M_2$, while the samples from the rehearsal buffer $B$ are passed through both models.
In this process, the mixed batch $\mathbf{b}^{\text{mix}}$ are converted into token sequences by the tokenizers $\mathcal{T}_{M_1}$ and $\mathcal{T}_{M_2}$ and then passed thorough the encoders $\phi_{M_1}$ and $\phi_{M_2}$, respectively.
The following feature distillation loss $\mathcal{L}_{\text{FD}}$ is calculated between the feature representations produced by $M_1$ and $M_2$, ensuring that $M_2$ does not deviate significantly from the feature representations learned in the first stage:
%
\begin{equation} 
\mathcal{L}_{\text{FD}} = || \phi_{M_2}(\mathbf{b}^{\text{mix}}) - \text{StopGrad}(\phi_{M_1}(\mathbf{b}^{\text{mix}})) ||_{2}^{2}, 
\end{equation}
where $\phi_{M_1}(\mathbf{b}^{\text{mix}})$ represents the feature representation of $\mathbf{b}^{\text{mix}}$ generated by the encoder $\phi_{M_1}$ from the first stage. 
$\phi_{M_2}(\mathbf{b}^{\text{mix}})$ is the feature representation of $\mathbf{b}^{\text{mix}}$ produced by the encoder $\phi_{M_2}$ during the third stage. 
\text{StopGrad} is applied to $\phi_{M_1}(\mathbf{b}^{\text{mix}})$ to ensure that no gradient flows back through $M_1$, keeping its parameters frozen during the training of $M_2$.
This feature distillation process helps model $M_2$ adapt to the new domain while preserving key knowledge from the first stage, ensuring better generalization across both domains, and balancing the knowledge from $D_1$ and $D_2$. 
Algorithm~\ref{alg1} summarizes the proposed CSSL method.
\par
By integrating an enhanced DER with a rehearsal-based strategy into the CSSL framework, the ViT encoder can efficiently capture the relationship between newly acquired data and previously learned knowledge across various medical domains. This approach minimizes data interference during pretraining and facilitates the learning of richer, more robust feature representations. Following the three-stage CSSL process, the pretrained ViT encoder can be fine-tuned on another labeled dataset for downstream tasks, such as classification.
\begin{algorithm}[t]
    \small
    \caption{Algorithm of the proposed CSSL method}
    \label{alg1}
    \begin{algorithmic}[1]
    \REQUIRE 
    $\{D_1, D_2\}$: two datasets with different domains, 
    $B$: rehearsal buffer, 
    $\mathcal{T}_{M_1}, \mathcal{T}_{M_2}$: tokenizers, 
    $\phi_{M_1}, \phi_{M_2}$: encoders,
    $\psi_{M_1}, \psi_{M_2}$: model-specific decoders,
    $Sample(\cdot)$: sampling operation
    \ENSURE
    $\phi_{M_2}$, $\mathcal{T}_{M_2}$ \\
    \textbf{Stage 1:}
    \STATE Training dataset $D \gets D_1$
    \STATE Update $\phi_{M_1}$, $\mathcal{T}_{M_1}$, and $\psi_{M_1}$ by $\mathcal{L}_{\text{MSE}}$
    \STATE $B \gets Sample(D_1)$
    \\
    \textbf{Stage 2:}
        \STATE Training dataset $D \gets D_2 \cup B$
        \FOR{ each interaction $i$ = 1 to I }
            \STATE Sample a batch of unlabeled data $x$ from $D$
            \IF{$x \in D_2$}
                \STATE Update $\phi_{M_2}$, $\mathcal{T}_{M_2}$, and $\psi_{M_2}$ by $L_{\text{MSE}}$
            \ELSE
                \STATE Update $\phi_{M_2}$ and $\mathcal{T}_{M_2}$ by $L_{\text{FD}}$
            \ENDIF
        \ENDFOR
    \end{algorithmic}
\end{algorithm}
\section{Experiments}
%
%
\subsection{Datasets and Settings}
We constructed two subsets of chest CT images from the J-MID database~\footnote{\url{https://www.radiology.jp/j-mid/}} based on mediastinal and lung window settings, and used them as $D_1$ and $D_2$ for pretraining.
The number of images $N_1$ and $N_2$ were 31,256 and 26,403, respectively. 
All of these images are grayscale and have a resolution of 512 $\times$ 512 pixels.
For fine-tuning and evaluation, we trained the model using the SARS-CoV-2 CT-Scan Dataset~\cite{cov-2_dataset} and validated it by performing a COVID-19 classification task.
The SARS-CoV-2 CT-Scan Dataset contains a total of 2,481 chest CT images labeled into two classes: COVID-19 positive and negative. 
Out of these, 1,986 images were used for fine-tuning and 495 images were used for evaluation.
In the pretraining of MAE, the masking ratio was set to $r = 0.75$, and the ViT-B~\cite{vit} was used as the encoder.
In each pretraining stage, a warm-up strategy was used during the first 40 epochs to gradually raise the learning rate from 0 to 0.00015, after which it was reduced to 0 in the following training using a cosine schedule.
In \textit{k}-means sampling, the parameters $\alpha$ and $\beta$, which determine the sampling ratio, were set to 0.01 and 0.05, respectively. 
Additionally, the number of clusters $K$ and the number of samples $T$ in the rehearsal buffer $B$ were set to 312 and 1,562, respectively. 
The parameters $\gamma_1$, $\gamma_2$, and $\gamma_3$, which determine the number of images to be acquired, were set to 6, 3, and 1, respectively.
The AdamW optimizer~\cite{adam} was used in the fine-tuning phase, and the learning rate was set to 0.00005.
We took SSL on the dataset for 300 epochs at each pretraining stage, followed by fine-tuning on the dataset for 80 epochs.
As evaluation metrics, two-class classification accuracy (ACC), the area under the receiver operator curve (AUC), and F1 score (F1) were used.
\par
The effectiveness of the learning order of domains in the proposed method was also verified by swapping $D_1$ and $D_2$ and conducting continual pretraining accordingly.
Furthermore, to examine how to consider the distance between data distributions of the two domains, we conducted an ablation study by varying the parameters $\gamma_1$, $\gamma_2$, and $\gamma_3$, which determine the number of images to be acquired in the rehearsal buffer (see Subsec 3.3 for details). 
\par
To evaluate the effectiveness of the proposed method (Ours), we compared it against several approaches: MedCoSS~\cite{medcoss}, the state-of-the-art CSSL method, MAE~\cite{MAE} with simultaneous pretraining on both $D_1$ and $D_2$, MAE pretrained only on $D_1$, and MAE pretrained only on $D_2$.
As a baseline method, we used MAE fine-tuned without self-supervised pretraining.
\subsection{Results and Discussion}
\begin{table}[t]
    \centering
    \small
    \caption{Experimental results of the proposed method and comparative methods on the SARS-CoV-2 CT-Scan Dataset.}
    \label{tab1}
    \begin{tabular}{lcccc}
    \hline
    Method & Domain & ACC & AUC & F1 \\\hline \hline
    Ours                        & \multirow{2}{*}{$D_1$ → $D_2$}   & \textbf{0.863} & \textbf{0.940} & \textbf{0.863} \\
    MedCoSS                     &                                   & 0.836 & 0.923 & 0.836 \\\hline\hline
    Ours                        & \multirow{2}{*}{$D_2$ → $D_1$}   & \textbf{0.800} & \textbf{0.900} & \textbf{0.798} \\
    MedCoSS                     &    & 0.756 & 0.837 & 0.755 \\\hline\hline
    \multirow{3}{*}{MAE}        &  $D_1$ + $D_2$  & 0.792 & 0.875 & 0.791 \\
                                &  $D_1$   & 0.717 & 0.803 & 0.713 \\
                                &  $D_2$   & 0.756 & 0.817 & 0.755 \\
    \hline \hline
    Baseline                    & None & 0.657 & 0.699 & 0.647 \\
    \hline
    \end{tabular}
\end{table}
\begin{figure}[t]
        \centering
        \includegraphics[width=8cm]{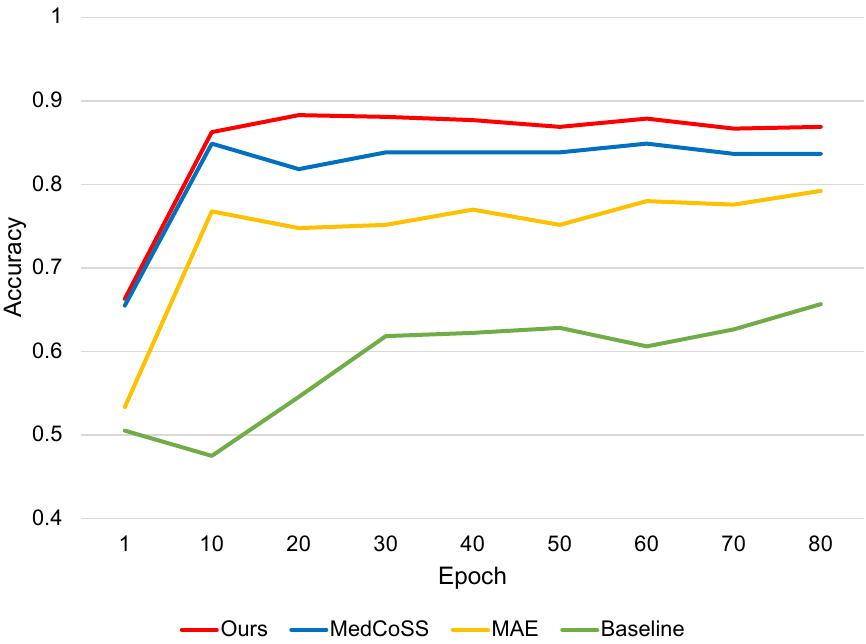}
        \caption{Classification accuracy across different epochs during fine-tuning. All methods use ViT-B model.}
        \label{fig3}
\end{figure}
From Table~\ref{tab1}, we can observe that the proposed method outperforms all the comparative methods.
From the three results of the MAE method, it was confirmed that pretraining with two domains improves accuracy compared to using a single domain.
Additionally, as the proposed method outperforms the MAE method that simultaneously pretrains on two domains in evaluation metrics, it was confirmed that the proposed method can reduce data interference between domains compared to pretraining on multiple domains simultaneously.
Furthermore, the experimental results show that when the model is pretrained on the domains in the same order, the proposed method outperforms MedCoSS in the evaluation metrics.
This confirms that the proposed method, which takes data distribution between stages into account, is effective in reducing data interference.
Furthermore, Fig.~\ref{fig3} shows that the proposed method achieves the fastest ACC convergence compared to all the comparative methods and consistently maintains the highest ACC at every number of epochs.
This indicates that maintaining not only diversity but also representativeness of sample data within the rehearsal buffer can effectively reduce data interference between stages.
\begin{table}[t]
    \small
    \centering
    \caption{
    Ablation study results of the proposed method on the SARS-CoV-2 CT-Scan Dataset. The best results are highlighted in bold, while the second-best are underlined.
    }
    \label{tab2}
    \begin{tabular}{ccccccc}
    \hline
    Method    & Domain & Ratio  & ACC   & AUC   & F1    \\\hline\hline
    \multirow{6}{*}{Ours} & \multirow{6}{*}{$D_1$ → $D_2$}  
                                        & 1:1:8     & 0.830 & 0.923 & 0.830 \\
                                        &            & 1:3:6     & 0.820 & 0.928 & 0.819 \\
                                        &            & 2:3:5     & 0.838 & 0.920 & 0.838 \\
                                        &            & \underline{5:3:2}     & \underline{0.842} & \underline{0.940} & \underline{0.842} \\
                                        &            & \textbf{6:3:1}     & \textbf{0.863} & \textbf{0.940} & \textbf{0.863} \\
                                        &            & 8:1:1     & 0.838 & 0.932 & 0.838 \\
    \hline                                        
    \multirow{6}{*}{Ours} & \multirow{6}{*}{$D_2$ → $D_1$}  
                                        & 1:1:8     & 0.669 & 0.712 & 0.668 \\
                                        &            & 1:3:6     & 0.778 & 0.856 & 0.777 \\
                                        &            & 2:3:5     & 0.756 & 0.838 & 0.755 \\
                                        &            & 5:3:2     & 0.760 & 0.862 & 0.756 \\  
                                        &            & 6:3:1     & 0.780 & 0.874 & 0.780 \\
                                        &            & 8:1:1     & 0.800 & 0.900 & 0.798 \\
    \hline
\end{tabular}
\end{table}
\subsection{Ablation Studies}
In the proposed method, we conducted an ablation study in which we varied the parameters $\gamma_1, \gamma_2,$ and $\gamma_3$, which determine the ratio of images acquired in the rehearsal buffer. 
This allows us to examine the optimal parameters that consider the differences in data distribution between stages to ensure diversity and representativeness within the rehearsal buffer.
The ablation results are shown in Table~\ref{tab2}.
In the proposed method, which performs continual pretraining in the order from $D_1$ to $D_2$, it is shown that the highest accuracy in the evaluation metrics is achieved when the parameters $\gamma_1, \gamma_2,$ and $\gamma_3$ are set to 6, 3, and 1, respectively.
Furthermore, it is observed that prioritizing the storage of samples that are closer in data distribution between domains in the rehearsal buffer, rather than those that are farther apart, tends to be more effective.
It was confirmed that focusing on representativeness when storing samples in the rehearsal buffer is effective. 
This is likely because the domains capture the same anatomical regions, leading to similar data distributions and a tendency for the data distributions between domains to be closer.
On the other hand, when $\gamma_1, \gamma_2,$ and $\gamma_3$ are set to 8, 1, and 1, respectively, the accuracy in the evaluation metrics decreases.
This suggests that it is preferable to maintain not only representativeness but also diversity.
\section{Conclusion}
In this paper, we presented a novel CSSL method that incorporates medical domain knowledge from chest CT images. By effectively capturing the differences in data distribution between two chest CT image domains and maintaining diversity and representativeness in the rehearsal buffer, our method enhances the representation learning process in CSSL. Additionally, we introduced a mixup strategy and feature distillation to further refine feature representations. Extensive experiments with chest CT images demonstrated that our approach outperforms state-of-the-art methods.
%
%
%
%
\bibliographystyle{IEEEbib}
\bibliography{refs}

\end{document}